\documentclass[10pt,twocolumn,letterpaper]{article}

\usepackage[pagenumbers]{cvpr} % To force page numbers, e.g. for an arXiv version

\usepackage{graphicx}
\usepackage{amsmath}
\usepackage{amssymb}
\usepackage{booktabs}
\usepackage{soul}
\usepackage{multirow}

\makeatletter
\@namedef{ver@everyshi.sty}{}
\makeatother

\usepackage{tikz}
\usepackage{pgfplots}
\pgfplotsset{compat=1.17}

\usepackage[pagebackref,breaklinks,colorlinks]{hyperref}

\usepackage[capitalize]{cleveref}
\crefname{section}{Sec.}{Secs.}
\Crefname{section}{Section}{Sections}
\Crefname{table}{Table}{Tables}
\crefname{table}{Tab.}{Tabs.}

\def\cvprPaperID{3040} % *** Enter the CVPR Paper ID here
\def\confName{CVPR}
\def\confYear{2023}

\begin{document}

\title{PETRv2: A Unified Framework for 3D Perception from Multi-Camera Images}

% \and

\author{Yingfei Liu  \qquad Junjie Yan  \qquad Fan Jia \qquad Shuailin Li \qquad Aqi Gao \\ Tiancai Wang\thanks{Corresponding author} \qquad Xiangyu Zhang \qquad Jian Sun \vspace{1mm}\\ MEGVII Technology\\
}

\maketitle

%%%%%%%%% ABSTRACT
\begin{abstract}  
In this paper, we propose PETRv2, a unified framework for 3D perception from multi-view images.
Based on PETR~\cite{liu2022petr}, PETRv2 explores the effectiveness of temporal modeling, which utilizes the temporal information of previous frames to boost 3D object detection. More specifically, we extend the 3D position embedding (3D PE) in PETR for temporal modeling. The 3D PE achieves the temporal alignment on object position of different frames. A feature-guided position encoder is further introduced to improve the data adaptability of 3D PE. To support for multi-task learning (e.g., BEV segmentation and 3D lane detection), PETRv2 provides a simple yet effective solution by introducing task-specific queries, which are initialized under different spaces. PETRv2 achieves state-of-the-art performance on 3D object detection, BEV segmentation and 3D lane detection. Detailed robustness analysis is also conducted on PETR framework. We hope PETRv2 can serve as a strong baseline for 3D perception. Code is available at \url{https://github.com/megvii-research/PETR}.
\end{abstract}

%%%%%%%%% BODY TEXT
%%%%%%%%%%%%%%%%%%%%%%%%%%%%%
\section{Introduction}
Recently, 3D perception from multi-camera images for autonomous driving system has drawn a great attention. The multi-camera 3D object detection methods can be divided into BEV-based~\cite{huang2021bevdet, huang2022bevdet4d} and DETR-based~\cite{wang2022detr3d, liu2022petr, li2022bevformer} approaches. 
BEV-based methods (e.g., BEVDet~\cite{huang2021bevdet}) explicitly transform the multi-view features into bird-eye-view (BEV) representation by LSS~\cite{philion2020lift}. Different from these BEV-based countparts, DETR-based approaches~\cite{wang2022detr3d} models each 3D object as an object query and achieve the end-to-end modeling with Hungarian algorithm~\cite{kuhn1955hungarian}. 
Among these methods, PETR~\cite{liu2022petr}, based on DETR~\cite{carion2020detr}, converts the multi-view 2D features to 3D position-aware features by adding the 3D position embedding (3D PE). The object query, initialized from 3D space, can directly perceive the 3D object information by interacting with the produced 3D position-aware features. In this paper, we aim to build a strong and unified framework by extending the PETR with temporal modeling and the support for multi-task learning.

For temporal modeling, the main problem is how to align the object position of different frames in 3D space. Existing works~\cite{huang2022bevdet4d, li2022bevformer} solved this problem from the perspective of feature alignment. For example, BEVDet4D~\cite{huang2022bevdet4d} explicitly aligns the BEV feature of previous frame with current frame by pose transformation. However, PETR implicitly encodes the 3D position into the 2D image features and fails to perform the explicit feature transformation.
% it seems not the case for PETR since there are no explicit position information in the produced 3D position-aware features.
% Since the keypoint of PETR is 3D position embedding (PE), we wonder if it is possible to achieve the temporal alignment from the perspective of 3D PE. 
Since PETR has demonstrated the effectiveness of 3D PE (encoding the 3D coordinates into 2D features) in 3D perception, we wonder if 3D PE still works on temporal alignment. 
In PETR, the meshgrid points of camera frustum space, shared for different views, are transformed to the 3D coordinates by camera parameters. The 3D coordinates are then input to a simple multi-layer perception (MLP) to generate the 3D PE. 
% Therefore, we can align the 3D coordinates of previous frames with current frame by pose transformation. 
In our practice, we find that PETR works well under temporal condition by simply aligning the 3D coordinates of previous frame with the current frame.

% For the joint learning with BEV segmentation, BEVFormer~\cite{li2022bevformer} provides a unified solution. It defines each point on BEV map as one BEV query. Thus, the BEV query can be employed for 3D object detection and BEV segmentation. However, the number of BEV query (e.g., >60,000) tends to be huge when the resolution of BEV map is relatively larger (e.g., $256 \times 256$). Such definition on object query is obviously not suitable for PETR due to the global attention employed in transformer decoder. In this paper, we design a simple and elegant solution for BEV segmentation. Inspired by the advanced instance segmentation methods~\cite{dong2021solq, shen2020dct,tian2020conditional,queryinst2021} where an instance mask within the bounding box is represented by a set of learnable parameters, we regard each patch of the BEV map as the mask within a bounding box and parameterize these BEV patches by different object queries. Despite object query for object detection (det query), we further introduce the so-called segmentation query (seg query) for BEV segmentation. The seg queries are initialized under BEV space and each seg query is responsible for segmenting specific patch. The updated seg queries from the transformer decoder are further used to predict the semantic map of corresponding patch. In this way, high-quality BEV segmentation can be achieved by simply adding a small number of (e.g., 256) seg queries.
For multi-task learning, BEVFormer~\cite{li2022bevformer} provides a unified solution. It defines each point on BEV map as one BEV query. Thus, the BEV query can be employed for 3D object detection and BEV segmentation. However, the number of BEV query (e.g., $>$60,000) tends to be huge when the resolution of BEV map is relatively larger (e.g., $256 \times 256$). Such definition on object query is obviously not suitable for PETR due to the global attention employed in transformer decoder. In this paper, we design a unified sparse-query solution for multi-task learning. For different tasks, we define  sparse task-specific queries under different spaces. For example, the lane queries for 3D lane detection are defined in 3D space with the style of anchor lane while seg queries for BEV segmentation are initialized under the BEV space. Those sparse task-specific queries are input to the same transformer decoder to update their representation and further injected into different task-specific heads to produce high-quality predictions.

Besides, we also improve the generation of 3D PE and provide a detailed robustness analysis on PETRv2. As mentioned above, 3D PE in PETR is generated based on the fixed meshgrid points in camera frustum space. All images from one camera view share the 3D PE, making 3D PE data-independent. In this paper, we further improve the original 3D PE by introducing a feature-guided position encoder (FPE). Concretely, the projected 2D features are firstly injected into a small MLP network and a Sigmoid layer to generate the attention weight, which is used to reweight the 3D PE in an element-wise manner. The improved 3D PE is data-dependent, providing the informative guidance for the query learning in transformer decoder. 
% It shows that 3D PE can implicitly achieve the temporal alignment in the 3D space. 
For comprehensive robustness analysis on PETRv2, we consider multiple noise cases, including the camera extrinsics noise, camera miss and time delay. 

To summarize, our contributions are:
\begin{itemize}
\item We study a conceptually simple extension of position embedding transformation to temporal representation learning. The temporal alignment can be achieved by the pose transformation on 3D PE. A feature-guided position encoder is further proposed to reweight the 3D PE with the guidance from 2D image features.
\item A simple yet effective solution is introduced for PETR to support the multi-task learning. BEV segmentation and 3D lane detection are supported by introducing task-specific queries.
% \item A feature-guided position encoder is further proposed to reweight the 3D PE with the guidance from 2D image features.
\item Experiments show that the proposed framework achieves state-of-the-art performance on both 3D object detection, BEV segmentation and 3D lane detection. Detailed robustness analysis is also provided for comprehensive evaluation on PETR framework.
\end{itemize}

%%%%%%%%%%%%%%%%%%%%%%%%%%%%%
\section{Related Work}
\subsection{Multi-View 3D Object Detection}
% Vision-based 3D object detection can be roughly divided into monocular and multi-view 3D detection. Many previous works are monocular methods and extend the 2D detectors to perform 3D object detection. Compared with FCOS, FCOS3D predict additional depth and other 3D cuboid parameters. PGD further models the uncertainty of depth to improve the performance of depth estimation. Besides, several works attempt to conduct the 3D object detection in 3D world space directly. Through camera intrinsic and extrinsic, OFT and CaDDN sample the image features and generate the voxel features to perform 3D detection. 
Previous works~\cite{chen2016monocular,mousavian20173d,kehl2017ssd,ku2019monocular,simonelli2019disentangling,jorgensen2019monocular,brazil2019m3d,wang2021fcos3d,wang2022pgd} perform 3D object detection mainly under the mono setting. Recently, 3D object detection based on multi-view images has attracted more attention. ImVoxelNet~\cite{rukhovich2022imvoxelnet} and BEVDet~\cite{huang2021bevdet} projected the multi-view image features into BEV representation. Then the 3D object detection can be performed using the methods from 3D point cloud, like~\cite{yin2021center}. DETR3D~\cite{wang2022detr3d} and PETR~\cite{liu2022petr} conduct the 3D object detection mainly inspired by the end-to-end DETR methods~\cite{carion2020detr,zhu2020deformable,meng2021conditional,liu2022dab}. The object queries are defined in 3D space and interact with the multi-view image features in transformer decoder. BEVFormer~\cite{li2022bevformer} further introduces the temporal information into vision-based 3D object detection. The spatial cross-attention is adopted to aggregate image features, while the temporal self-attention is used to fuse the history BEV features. BEVDet4D~\cite{huang2022bevdet4d} extends the BEVDet~\cite{huang2021bevdet} by the temporal modeling and achieves good speed estimation. Both BEVFormer~\cite{li2022bevformer} and BEVDet4D~\cite{huang2022bevdet4d} align the multi-frame features in BEV space. Different from them, we extend the temporal version from PETR and achieve the temporal alignment from the perspective of 3D position embedding (3D PE). 
% Considering that the transformation of BEV features is complicated and lossy, this paper try to do the alignment in the image view.

\subsection{BEV Segmentation}
BEV segmentation focus on the perception in the BEV view. It takes the multi-view images as input and rasterizes output onto a map view.
VPN~\cite{pan2020cross} proposes a view parsing network under the simulated environments and then transfers it to real-world environments to perform cross-view semantic segmentation.
% There are also some multi-task studies in 3D computer vision community. 
LSS~\cite{philion2020lift} transforms the 2D features into 3D space by implicit estimation of depth and employs different heads for BEV segmentation and planning. 
M$^{2}$BEV~\cite{xie2022m} further uses the camera parameters to project the features extracted from backbone to the 3D ego-car coordinate to generate the BEV representation. Then multi-task heads are used for 3D detection and segmentation.
BEVFormer~\cite{li2022bevformer} generates the BEV features from multi-camera inputs by interacting the predefined grid-shaped BEV queries with the 2D image features.
CVT~\cite{zhou2022cross} uses cross-view transformer to learn geometric transformation implicitly.
HDMapNet~\cite{li2021hdmapnet} transforms multi-view images to the BEV view and produces a vectorized local semantic map.  
BEVSegFormer~\cite{peng2022bevsegformer} proposes multi-camera
deformable attention to construct semantic map. 

\subsection{3D Lane Detection}
BEV segmentation can reconstruct the elements of local map. However, it fails to model the spatial association between different instances. Recently, the 3D lane detection task has attracted more and more attention. 3D-LaneNet~\cite{garnett20193d} is the first method that makes the 3D lane prediction. It uses inverse perspective mapping (IPM) to transform feature from front view to BEV. Gen-LaneNet~\cite{guo2020gen} introduces a new anchor lane representation to align the perspective anchor representation and BEV feature. Persformer~\cite{chen2022persformer} employs the deformable attention to generate BEV features by attending local context around reference points.    
CurveFormer~\cite{bai2022curveformer} introduces a curve cross-attention module to compute the similarities between curve queries and image features. It employs deformable attention to obtain the image features corresponding to the reference points.
% However, all of the above methods use multi-view images to construct feature maps from the BEV perspective, and then use different heads to perform multi-tasks. The construction of such BEV feature maps consumes a lot of computing resources, so it is difficult to construct high resolution BEV features. That may lead to the loss of information in the final generated BEV features. Our method uses different 3D queries to interact with multi-view features, which can effectively obtain more spatial features.

%%%%%%%%%%%%%%%%%%%%%%%%%%%%%
\section{Method}

\begin{figure*}[t]
	\centering  
	\includegraphics[width=0.85\linewidth]{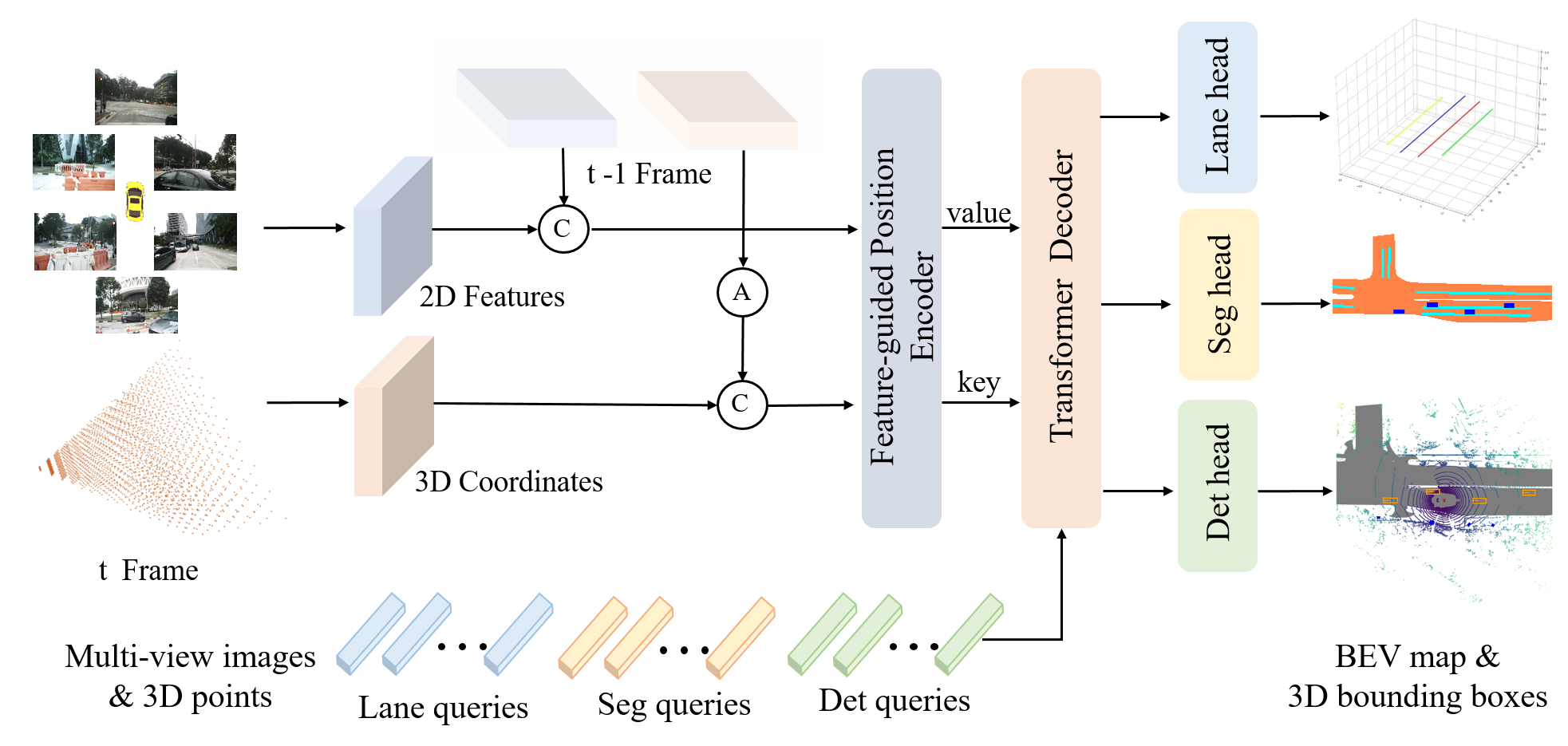}
	\caption{The paradigm of the proposed PETRv2. The 2D features are extracted by the backbone network from the multi-view images and the 3D coordinates are generated following the same way as PETR~\cite{liu2022petr}. To achieve the temporal alignment, the 3D coordinates in PETR of previous frame $t-1$ are firstly transformed through pose transformation. Then 2D image features and 3D coordinates of two frames are concatenated together and injected to feature-guided position encoder to generate the key and value components for the transformer decoder. The detection, segmentation and lane queries, initialized under different spaces, interact with the key and value components in transformer decoder. The updated queries are further used to predict the 3D bounding boxes, BEV segmentation map and the 3D lanes with task-specific heads. $\textcircled{\scriptsize A}$ is 3D coordinates alignment from  frame $t-1$ to  frame $t$. $\textcircled{\scriptsize C}$ is concatenation operation along the batch axis.}  
	\label{architecture}
\end{figure*}

\subsection{Overall Architecture}
As illustrated in Fig.~\ref{architecture}, the overall architecture of PETRv2 is built upon the PETR~\cite{liu2022petr} and extended with temporal modeling and BEV segmentation. The 2D image features are extracted from multi-view images with the 2D backbone (e.g., ResNet-50), and the 3D coordinates are generated from camera frustum space as described in PETR~\cite{liu2022petr}. Considering the ego motion, 3D coordinates of the previous frame $t-1$ are first transformed into the coordinate system of current frame $t$ through the pose transformation. Then, the 2D features and 3D coordinates of adjacent frames are respectively concatenated together and input to the feature-guided position encoder (FPE). After that, the FPE is employed to generate the key and value components for the transformer decoder.
% the 3D PE in a data-dependent way. The 3D PE is feature-guided, which is the key of position alignment among multi-frames.
Further, task-specific queries including the detection queries (det queries) and segmentation queries (seg queries), which are initialized from different spaces, are fed into the transformer decoder and interact with multi-view image features. Lastly, the updated queries are input to the task-specific heads for final prediction. 
% $\textcircled{\scriptsize A}$ is the coordinate transformation operation.

\subsection{Temporal Modeling}

% \begin{figure*}[t]
% 	\centering  
% 	\begin{subfigure}{.45\textwidth}
% 			\centering
% 			\includegraphics[width=\textwidth]{car.png}
% 			\caption{}
% 		\end{subfigure}
% 	\begin{subfigure}{.45\textwidth}
% 			\centering
% 			\includegraphics[width=\textwidth]{dse.png}
% 			\caption{}
% 		\end{subfigure}
% 	\caption{(a) The illustration of the coordinate system transformation from frame $t-1$ to frame $t$.
% % 	Different 3D points are sampled for adjacent frames to produce the 3D position embedding (PE).
% 	(b) Architecture of feature-guided position encoder. Different from PETR~\cite{liu2022petr}, 3D PE in PETRv2 is generated in a data-dependent way.}
% 	\label{dpe module}
% 	\vspace{-1.0em}
% \end{figure*}

PETR~\cite{liu2022petr} leverages image features and projected 3D points to generate implicit 3D features for multi-view 3D detection. In this section, we extend it with the temporal modeling, which is realized by a 3D coordinates alignment (CA) for better localization and speed estimation. 

\begin{figure}[h]
	\centering  
	\includegraphics[width=0.85\linewidth]{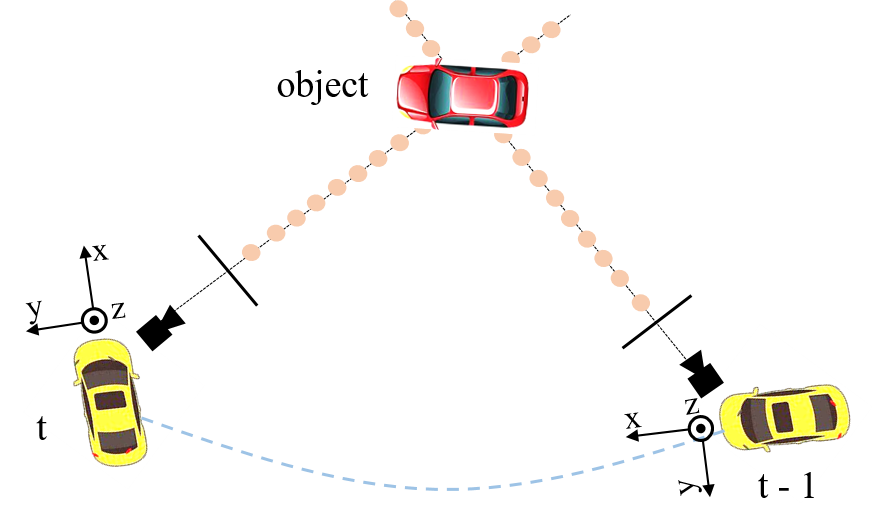}
	\caption{The illustration of the coordinate system transformation from frame $t-1$ to frame $t$.
    }  
	\label{coordinate}
\end{figure}

\noindent \textbf{3D Coordinates Alignment} 
% For clarity, we first denote some coordinate space, camera coordinate as $cam(t)$, lidar coordinate as  $lidar(t)$, and ego coordinate as $ego(t)$ at frame $t$. What's more, global coordinates as $global$.
% We use $T^{dst}_{src}$ as the transform matrix between the source coordinate space and the target coordinate space.
% We use lidar space  $lidar(t)$ as the default 3D space for multi-view camera 3D feature generation. The 3D points $P^{lidar(t)}_i(t)$ projected from $i$-th camera frustum space can be formulated as:
% \begin{equation}
% \begin{split}
%     P^{lidar(t)}_i(t)) = T^{lidar(t)}_{cam_i(t)} K^{-1}_{i} P^{m}(t) \\
% \end{split}
% \end{equation}
% When fusion multi-frames, for example, $t-1$ and $t$, we align the 3D points in the frame $t-1$ from $lidar(t-1)$ coordinate space to $lidar(t)$ as follows:
% \begin{equation}
% \begin{split}
%     P^{lidar(t)}_i(t-1) = T^{lidar(t)}_{lidar(t-1)} P^{lidar(t-1)}_i(t-1) 
% \end{split}
% \end{equation}
% \begin{equation}
%     \begin{split}
%         T^{lidar(t)}_{lidar(t-1)} &= T^{lidar(t)}_{global} {T^{lidar(t-1)}_{global}}^{-1} \\
%         &= T^{lidar(t)}_{ego(t)} T^{ego(t)}_{global} {T^{ego(t-1)}_{global}}^{-1} {T^{lidar(t-1)}_{ego(t-1)}}^{-1} \\
%     \end{split}
% \end{equation}
The temporal alignment is to transform the 3D coordinates of frame $t-1$ to the coordinate system of frame $t$ (see Fig.~\ref{coordinate}). For clarity, we first denote some coordinate systems: camera coordinate as $c(t)$, lidar coordinate as  $l(t)$, and ego coordinate as $e(t)$ at frame $t$. What's more, global coordinates as $g$.
We define $T^{dst}_{src}$ as the transformation matrix from the source coordinate system to the target coordinate system.

\begin{figure*}[t]
	\centering  
	\includegraphics[height=3.7cm,width=15cm]{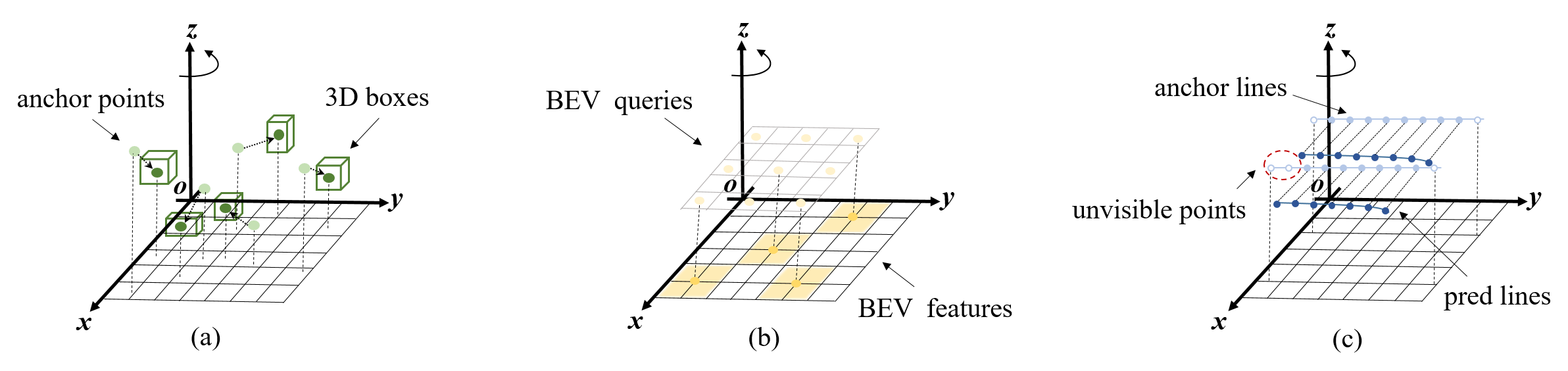}
    \vspace{-10pt}
	\caption{
	 The definition of three kinds of queries for multi-task learning. The det query is defined in the whole 3D space while the seg query is initialized under the BEV space. The lane query is defined with the anchor line, which is constructed with 300 anchor points.  
    }  
	\label{seg}
\end{figure*}

We use $l(t)$ as the default 3D space for multi-view camera 3D position-aware feature generation. The 3D points $P^{l(t)}_i(t)$ projected from $i$-th camera can be formulated as:
\begin{equation}
\begin{split}
    P^{l(t)}_i(t) = T^{l(t)}_{c_i(t)} K^{-1}_{i} P^{m}(t) \\
\end{split}
\end{equation}
where $P^{m}(t)$ is the points set in the meshgrid of camera frustum space at frame $t$. $K_i\in R^{4 \times 4}$ is the camera intrinsic matrix of the $i$-th camera.
Given the auxiliary frame $t-1$, we align the coordinates of 3D points from frame $t-1$ to frame $t$:
\begin{equation}
\begin{split}
    P^{l(t)}_i(t-1) = T^{l(t)}_{l(t-1)} P^{l(t-1)}_i(t-1) 
\end{split}
\end{equation}
With $global$ coordinate space acting as a bridge between frame $t-1$ and frame $t$, $T^{l(t)}_{l(t-1)}$ can be easily calculated:
\begin{equation}
    \begin{split}
        % T^{l(t)}_{l(t-1)} &= T^{l(t)}_{g} {T^{l(t-1)}_{global}}^{-1} \\
        % &= T^{l(t)}_{e(t)} T^{e(t)}_{g} {T^{e(t-1)}_{global}}^{-1} {T^{l(t-1)}_{e(t-1)}}^{-1} \\
        T^{l(t)}_{l(t-1)} &= T^{l(t)}_{e(t)} T^{e(t)}_{g} {T^{e(t-1)}_{g}}^{-1} {T^{l(t-1)}_{e(t-1)}}^{-1} \\
    \end{split}
\end{equation}
% We latter encode 2D image features and 3D points $[P^{l(t)}_i(t-1), P^{l(t)}_i(t)]$ together to obtain 3D features for each image separately.
The aligned point sets $[P^{l(t)}_i(t-1), P^{l(t)}_i(t)]$ are used to generate the 3D PE, as described in Sec.~\ref{sec:fpe}. 

\subsection{Multi-task Learning}
In this section, we aim to equip PETR~\cite{liu2022petr} with seg queries and lane queries to support high-quality BEV segmentation and 3D Lane detection.
%Most studies on HD map segmentation task first construct BEV feature map as the input of segmentation head. This method maps features to BEV perspective and eliminates height information, which will result in feature loss from multiple views. 
% In this section, we introduce the segmentation queries to achieve high-quality BEV segmentation. Because BEV segmentation results are distributed over the entire BEV space, the initialization of the query is fixed in the BEV space. We use multi-view feature with position embedding to interact with the predefined grid-shaped queries in transformer decoder instead of generating BEV feature map. Then the updated predefined grid-shaped queries are further used to predict a specific part of the HD Map.

\noindent \textbf{BEV Segmentation} 
% In this section, we aim to equip the PETR~\cite{liu2022petr} with seg queries to support high-quality BEV segmentation.
A high-resolution BEV map can be partitioned into a small number of patches. 
% then we can segment regions based on the patch level rather than pixel level. 
% Inspired by SOLQ~\cite{dong2021solq} where an instance mask within the bounding box can be represented by a set of learnable parameters, we regard each patch of the BEV map as the region within a bounding box and parameterize the BEV patches by object queries.
We introduce the seg query for BEV segmentation and each seg query corresponds to a specific patch (e.g., top-left $25 \times 25$ pixels of the BEV map). As shown in Fig.~\ref{seg} (b), the seg queries are initialized with fixed anchor points in BEV space, similar to the generation of detection query (det query) in PETR. These anchor points are then projected into the seg queries by a simple MLP with two linear layers. After that, the seg queries are input to the transformer decoder and interact with the image features. For the transformer decoder, we use the same framework as detection task. Then the updated seg queries are finally fed into the segmentation head, similar to the decoder in CVT~\cite{zhou2022cross}, to predict the final segmentation results. We use focal loss to supervise the predictions of each category separately.

\noindent \textbf{3D Lane Detection}
We add lane queries on PETR to support 3D lane detection (see Fig.~\ref{seg} (c)). We define the 3D anchor lanes, each of which is represented as an ordered set of 3D coordinates: $l=\{(x_1,y_1,z_1,),(x_2,y_2,z_2), \cdots , (x_n,y_n,z_n) \}$, where $n$ is the number of the sample points of each lane. In order to improve the prediction ability for 3D lanes, we use a fixed sampling point set uniformly sampled along the Y-axis, similar to Persformer~\cite{chen2022persformer}. 
% the anchor lanes consist of the points that uniformly placed longitudinal along Y-axis.
Different from Persformer, our anchor lanes are parallel to the Y-axis while the Persformer predefines different slopes for each anchor line. The updated lane queries from transformer decoder are used to predict the 3D lane instances. The 3D lane head predicts the lane class $C$ as well as the relative offset $(\Delta x,\Delta z)$ along x-axis and z-axis compared to the anchor lanes. Since the length of 3D lane is not fixed, we also predict the visibility vector $T_{vis}$ of size $n$ to control the start and end points of the lane. We use focal loss to supervise the predictions of the lane category and visibility. We also use  $L1$ loss to supervise the predictions of the offset. 
% $(C, \{(\Delta X^i,\Delta Z^i, T_{vis}^i )\}_{i=0}^n)$ for each lane query. Where C is category of the 3D lane, $\Delta X^i$ and $\Delta Z^i$ is the offset along x-axis and z-axis of each location $i$ in an anchor lane. 
% Since the length of 3D lane is not fixed, we predict the visibility vector $T_{vis}$ of size $n$ to control the start and end points of the lane. 

\subsection{Feature-guided Position Encoder}
\label{sec:fpe}
\begin{figure}[h]
	\centering  
	\includegraphics[width=0.85\linewidth]{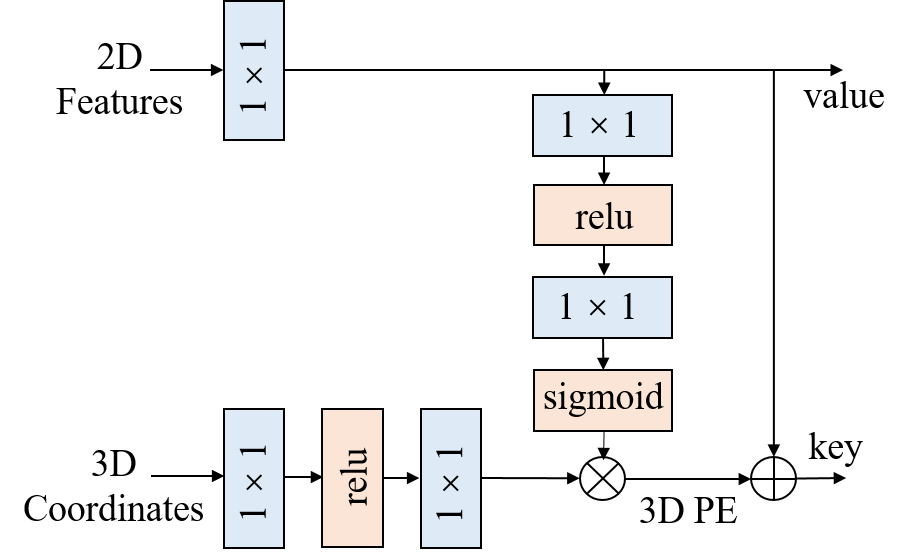}
	\caption{Architecture of feature-guided position encoder. Different from PETR~\cite{liu2022petr}, 3D PE in PETRv2 is generated in a data-dependent way and guided by the image features.
    }  
	\label{dse}
\end{figure}

PETR transforms the 3D coordinates into 3D position embedding (3D PE).
The generation of 3D position embedding can be formulated as:
\begin{equation}\label{petr_pe}
PE^{3d}_i(t) = \psi(P^{l(t)}_i(t))
\end{equation}
where $\psi(.)$ is a simple multi-layer perception (MLP). 
% However, the origin 3D position embedding of two frames are misaligned. 
The 3D PE in PETR is independent with the input image. We argue that the 3D PE should be driven by the 2D features since the image feature can provide some informative guidance (e.g., depth).
% As shown in Fig.~\ref{dpe module}(a), influenced by ego motion, different 3D points are sampled from two frames for a static object. Even if the coordinates system of frame $t-1$ has been aligned to frame $t$, the points sampled from two adjacent frames for the same static objects are still different. The intuitive solution is to sample points at the precise object location. However, the depth estimation is difficult and inaccurate. 
In this paper, we propose a feature-guided position encoder, which implicitly introduces vision prior. 
The generation of feature-guided 3D position embedding can be formulated as:
\begin{equation}\label{fpe}
\begin{split}
PE^{3d}_i(t) &= \xi(F_i(t)) \ast \psi(P^{l(t)}_i(t)) \\
% PE^{3d}_i(t-1) &= \xi(F_i(t-1)) \ast \psi(P^{l(t)}_i(t-1))
\end{split}
\end{equation}
where $\xi$ is also a small MLP network. $F_{i}(t)$ is the 2D image features of the $i$-th camera. As illustrated in Fig.~\ref{dse}, the 2D image features projected by a $1\times1$ convolution are fed into a small MLP network $\xi$ and Sigmoid function to obtain the attention weights. The 3D coordinates are transformed by another MLP network $\psi$ and multiplied with the attention weights to generate the 3D PE. The 3D PE is added with 2D features to obtain the key value for transformer decoder. The projected 2D features are used as the value component for transformer decoder.
% We simply concatenate multi-frame features for further prediction. 

\subsection{Robustness Analysis}
% Outline:

\begin{figure}[h]
	\centering  
    \includegraphics[width=1.0\linewidth]{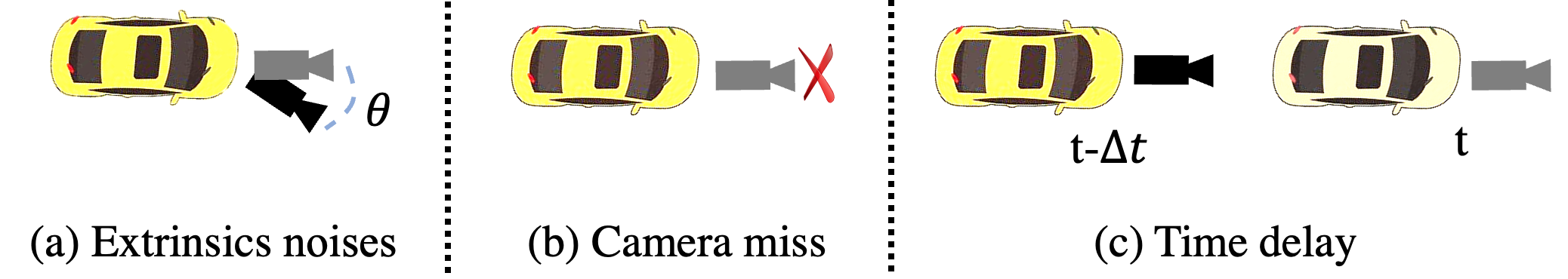}
	\caption{We analyze the system robustness of PETR series under three simulated sensor errors: (a) extrinsics noise, (b) camera miss and (c) camera time delay.}  
	\label{fig:robust}
\end{figure}

% 1. background: 
% - extrinsic noise are common in real world, such as ...
% - several types: extrinsic noise, camera image miss, time delay 
% - it is significant to present the performance on different conditions...
Though recently there are lots of works on autonomous driving systems, only a few works~\cite{philion2020lift,li2022bevformer} explore the robustness of proposed methods.
LSS~\cite{philion2020lift} presents the performance under extrinsics noises and camera dropout at test time. 
Similarly, BEVFormer~\cite{li2022bevformer} demonstrates the robustness of model variants to camera extrinsics.
In practice, there are diverse sensor errors and system biases, and it is important to validate the effect of these circumstances due to the high requirements of safety and reliability. 
We aim to give an extensive study of our method under different conditions. As shown in Fig.~\ref{fig:robust}, 
we focus on three common types of sensor errors as follows:

\noindent \textbf{Extrinsics noise:} Extrinsics noises are very common in reality, such as the camera shake caused by a car bump or camera offset by the environmental forces. In these cases, extrisics provided by the system is not that accurate and the perception results will be affected.

\noindent \textbf{Camera miss:} Camera image miss occurs when one camera breaks down or is occluded. Multiview images provide panoramic visual information, yet the possibility exists that one of them is absent in the real world. It is necessary to evaluate the importance of these images so as to prepare the strategy of sensor redundancy in advance.

\noindent \textbf{Camera time delay:} Camera time delay is also a challenge due to the camera exposure time, especially in night. The long exposure time causes the system is fed with images from the previous time, and brings the significant output offsets.

\section{Experiments}

\subsection{Datasets and Metrics}
We evaluate our approach on nuScenes benchmark~\cite{caesar2020nuscenes} and OpenLane benchmark~\cite{chen2022persformer}. NuScenes~\cite{caesar2020nuscenes} is a large-scale multi-task dataset covering 3D object detection, BEV segmentation, 3D object tracking, etc. 
The dataset is officially divided into training/validation/testing sets with 700/150/150 scenes, respectively.
We mainly focus on two sub-tasks: 3D object detection and BEV segmentation. 
We also conduct the 3D lane detection experiments on OpenLane benchmark~\cite{chen2022persformer}. Openlane ~\cite{chen2022persformer} is a large-scale real world 3D lane dataset. It has 200K frames and over 880K carefully annotated lanes and covers a wide range of lane types using 14 lane categories.

For 3D object detection, each scene has 20s video frames and is annotated around 40 key frames. We report the official evaluation metrics including nuScenes Detection Score (NDS), mean Average Precision (mAP), and five True Positive (TP) metrics: mean Average Translation Error (mATE), mean Average Scale Error (mASE), mean Average Orientation Error(mAOE), mean Average Velocity Error(mAVE), mean Average Attribute Error(mAAE). NDS is a comprehensive indicator to evaluate the detection performance.
% \begin{equation}\label{eq3}
% NDS = \frac{1}{10}[5 mAP + \sum_{mTP\in TP}(1 - min(1, mTP))]  
% \end{equation}

For BEV segmentation, we follow LSS \cite{philion2020lift} and use IoU score as the metric. The ground-truth includes three different categories: Driveable area, Lane and Vehicle. The lane category is formed by two map layers: lane-Divider and Road-Divider. For Vehicle segmentation, we obtain the BEV ground truth by projecting 3D bounding boxes into the BEV plane\cite{philion2020lift}. The Vehicle segmentation ground truth refers to all bounding boxes of meta-category Vehicle, which contains bicycle, bus, car, construction, motorcycle, trailer and truck.

For 3D lane detection, we follow Persformer ~\cite{chen2022persformer} using F1-Score and category accuracy as the metrics.
When $75\%$ points of a predicted lane instance have the point-wise euclidean distance less than 1.5 meters, the lane instance is considered to be correctly predicted.
We also report X error near, X error far, Z error near, Z error far to evaluate the models. These four metrics are used to evaluate the average error of the results in specified ranges.

\begin{table*}[t!]
\begin{center}
\caption{Comparison of recent works on the nuScenes val set. The results of FCOS3D and PGD are fine-tuned and tested with test time augmentation. The DETR3D, BEVDet and PETR are trained with CBGS~\cite{zhu2019class}. $\dagger$ is initialized from a FCOS3D backbone.
}
\label{table:1}
\begin{tabular}{l|cc|ccccccc}
\hline
% \noalign{\smallskip}
Methods & Backbone & Size  & NDS$\uparrow$ & mAP$\uparrow$ & mATE$\downarrow$ & mASE$\downarrow$ & mAOE$\downarrow$ & mAVE$\downarrow$ & mAAE$\downarrow$ \\

% \noalign{\smallskip}
\hline
% \noalign{\smallskip}
 CenterNet~\cite{zhou2019objects}&DLA  &  - &0.328 &0.306 &0.716 &0.264 &0.609 &1.426 &0.658   \\
 FCOS3D~\cite{wang2021fcos3d}  &Res-101  & 1600$\times$900  &0.415 &0.343 &0.725 &0.263 &0.422 &1.292 &\textbf{0.153} \\
 PGD~\cite{wang2022pgd} &Res-101 & 1600$\times$900  &0.428 &0.369  &0.683 &0.260 &0.439 &1.268 &0.185  \\
BEVDet~\cite{huang2021bevdet} &Swin-T & 1408$\times$512  &0.417 &0.349 &0.637 &0.269 &0.490 &0.914 &0.268 \\
DETR3D$\dagger$~\cite{wang2022detr3d} &Res-101  & 1600$\times$900  &0.434 &0.349 &0.716 &0.268 &0.379 &0.842 &0.200 \\
PETR$\dagger$~\cite{liu2022petr}  &Res-101 & 1600$\times$900 &0.442 &0.370 &0.711 &0.267 &0.383 &  0.865 & 0.201 \\
\hline
BEVFormer$\dagger$~\cite{li2022bevformer}  &Res-101 & 1600$\times$900 &0.517 &0.416 &0.673 &0.274 &0.372 &0.394 &0.198 \\
BEVDet4D~\cite{huang2022bevdet4d}  &Swin-B & 1600$\times$640 &0.515 &0.396 &\textbf{0.619} &\textbf{0.260} &0.361 &0.399 &0.189 \\
PETRv2 &Res-50 & 800$\times$320 &0.456 &0.350 &0.726 &0.277 &0.505 &0.503 &0.181 \\
PETRv2 &Res-50 & 1600$\times$640 &0.494 &0.398 &0.690 &0.273 &0.467 &0.424 &0.195 \\
PETRv2$\dagger$  &Res-101 & 800$\times$320 &0.489 &0.375 &0.677 &0.271 &0.414 &0.435 &0.192 \\
PETRv2$\dagger$  &Res-101 & 1600$\times$640 &\textbf{0.524} &\textbf{0.421} &0.681 &0.267 &\textbf{0.357} &\textbf{0.377} &0.186 \\
\hline
\end{tabular}
\end{center}

\end{table*}
\setlength{\tabcolsep}{1pt}

\subsection{Implementation Details}
In our implementation, ResNet~\cite{he2016resnet}, VoVNetV2~\cite{lee2020centermask} and EfficientNet ~\cite{tan2019efficientnet} are employed as the backbone for feature extraction. The P4 feature (merging the C4 and C5 features from backbone) with 1/16 input resolution is used as the 2D feature. The generation of 3D coordinates is consistent with PETR~\cite{liu2022petr}. Following BEVDet4D~\cite{huang2022bevdet4d}, we randomly sample a frame as previous frame ranging from [$3T$, $27T$] during training, and sample the frame at $15T$ during inference.
$T (\approx 0.083)$ is the time interval between two sweep frames. Our model is trained using AdamW~\cite{loshchilov2017decoupled} optimizer with a weight decay of 0.01. The learning rate is initialized with $2.0\times10^{-4}$ and decayed with cosine annealing policy~\cite{loshchilov2016sgdr}. All the experiments are trained for 24 epochs (2$\times$ schedule) on 8 Tesla A100 GPUs with a total batch size of 8 except for the ablation study. No test time augmentation methods are used during inference.

For 3D object detection, we perform experiments with 1500 det queries on nuScenes test dataset. Following the settings in PETR~\cite{liu2022petr}, we initialize a set of learnable anchor points in 3D world space, and generate these queries through a small MLP network. Similar to FCOS3D~\cite{wang2021fcos3d}, we add extra disentangled layers for regression targets. 
We extend query denoise of DN-DETR~\cite{li2022dn} to accelerate convergence of 3D object detection. For each ground-truth 3D box, the center is shifted by a random noise less than ($w$/2, $l$/2, $h$/2), where ($w$, $l$, $h$) is the size of object.
We also adopt the focal loss~\cite{lin2017focal} for classification and $L1$ loss for 3D bounding box regression. The Hungarian algorithm~\cite{kuhn1955hungarian} is used for label assignment between ground-truths and predictions. For BEV segmentation, we follow the settings in \cite{philion2020lift}. We use the map layers provided by the nuScenes dataset to generate the $200 \times 200$ BEV map ground truth. We set the patch size to $25 \times 25$ and $625$ seg queries are used to predict the final BEV segmentation result. For 3D lane detection, we follow the settings in ~\cite{chen2022persformer}.
The input size of images is $360 \times 480$. We use $100$ lane queries to predict the 3D lanes.  
We set the number of points in each anchor lane to 10 and the prediction range is $[3m,103m]$ on Y-axis and $[-10m,10m]$ on X-axis. The distance is calculated at several fixed positions along the Y-axis: [5, 10, 15, 20, 30, 40, 50, 60, 80, 100] for 3D anchor lanes.  

% setting
To simulate extrinsic noises and evaluate the effect, we choose to randomly apply 3D rotation to camera extrinsics. 3D rotation is very common and typical in real scenarios, and we ignore other noisy patterns such as translation to avoid multi-variable interference.
Specifically, we randomly choose one from multiple cameras to apply 3D rotation. Denoting $\alpha, \beta, \gamma$ as angles (in degree) along $X, Y, Z$ axes respectively, we investigate in several rotation settings with maximum amplitudes $\alpha_{max}, \beta_{max}, \gamma_{max} \in \{2, 4, 6, 8\}$, where $\alpha_{max}=2$ means that $\alpha$ is uniformly sampled from $[-2, 2]$, for example. In experiment, we use $R_{max}=M$ to denote $\alpha_{max} = \beta_{max} = \gamma_{max} = M$.

\begin{table*}[t!]
\begin{center}
\caption{Comparison of recent works on the nuScenes test set. $\ast$ are trained with external data.  $\ddagger$ is test time augmentation. “ms ” indicates using the resolution of $800\times320$ and $1600\times640$ as the inputs.}
\label{table:2}
\setlength{\tabcolsep}{5pt}
\begin{tabular}{l|c|ccccccc}
\hline
% \noalign{\smallskip}
Methods & Backbone & NDS$\uparrow$ & mAP$\uparrow$ & mATE$\downarrow$ & mASE$\downarrow$ & mAOE$\downarrow$ & mAVE$\downarrow$ & mAAE$\downarrow$ \\

% \noalign{\smallskip}
\hline
% \noalign{\smallskip}
 CenterNet~\cite{zhou2019objects} & DLA &0.400 &0.338 &0.658 &0.255 &0.629 &1.629 &0.142  \\
 FCOS3D$\ddagger$~\cite{wang2021fcos3d} & Res-101 &0.428 &0.358 &0.690 &0.249 &0.452 &1.434 &0.124  \\
 PGD$\ddagger$~\cite{wang2022pgd} & Res-101 &0.448 &0.386 &0.626 &0.245 &0.451 &1.509 &0.127  \\
 DD3D$\ast\ddagger$~\cite{park2021dd3d} & V2-99 &0.477 &0.418  &0.572 &0.249 &0.368 &1.014 &0.124  \\
 DETR3D$\ast$~\cite{wang2022detr3d} & V2-99 &0.479 &0.412  &0.641 &0.255 &0.394 &0.845 &0.133  \\
 BEVDet~\cite{huang2021bevdet} & Swin-S &0.463 &0.398 &0.556 &0.239 &0.414 &1.010 &0.153  \\
 BEVDet$\ast$~\cite{huang2021bevdet} & V2-99 &0.488 &0.424 &0.524 &0.242 &0.373 &0.950 &0.148  \\
 M$^{2}$BEV~\cite{xie2022m} & X-101&0.474 &0.429 &0.583 &0.254 & 0.376 &1.053 &0.190 \\
 PETR$\ast$~\cite{liu2022petr} & V2-99 &0.504 &0.441 &0.593 &0.249 &0.383 &0.808 &0.132  \\
\hline
BEVFormer~\cite{li2022bevformer} & Res-101 &0.535 &0.445 &0.631 &0.257 &0.405 &0.435 &0.143  \\
BEVFormer$\ast$~\cite{li2022bevformer} & V2-99 &0.569 &0.481 &0.582 &0.256 &0.375 &0.378 &0.126  \\
BEVDet4D$\ddagger$~\cite{huang2022bevdet4d} & Swin-B &0.569 &0.451 &\textbf{0.511} &\textbf{0.241} &0.386 &\textbf{0.301} &0.121  \\
% PETRv2$\ast$ & V2-99 &0.574 &0.483 &0.580 &0.247 &0.359 &0.369 &0.117  \\
PETRv2 & Res-101 &0.553 &0.456 &0.601 &0.249 &0.391 &0.382 &0.123  \\
% PETRv2$\ast$ & V2-99 &\textbf{0.582} &\textbf{0.490} &0.561 &0.243 &\textbf{0.361} &0.343 &\textbf{0.120}  \\
PETRv2$\ast$ & V2-99 &0.582 &0.490 &0.561 &0.243 &0.361 &0.343 &0.120  \\
PETRv2$\ast$ ms & V2-99 &\textbf{0.591} &\textbf{0.508} &0.543 &\textbf{0.241} &\textbf{0.360} &0.367 &\textbf{0.118}  \\

\hline
\end{tabular}
\end{center}
% \vspace{-1.5em}
\end{table*}

\begin{table}[t]
\begin{center}
\caption{Comparison of recent BEV segmentation works on the nuScenes val set. $\ast$ are trained with external data. The performance of M$^{2}$BEV is reported with X-101~\cite{xie2017aggregated} backbone.}
\label{seg_compare}
\setlength{\tabcolsep}{5pt}
\begin{tabular}{l|c|ccc}
\hline
% \noalign{\smallskip}
Methods & Backbone  & Drive & Lane & Vehicle  \\
\hline
Lift-Splat~\cite{philion2020lift} & Res-101&0.729 &0.200 &0.321   \\
FIERY~\cite{hu2021fiery} & Res-101& - &- &0.382 \\
M$^{2}$BEV~\cite{xie2022m} & X-101&0.759 &0.380 &- \\
BEVFormer~\cite{li2022bevformer} & Res-101& 0.801 &0.257 &0.448 \\
\hline

PETRv2 & Res-101  &0.833 &0.448 &0.434   \\
%PETR-segonly & Res-101   &0.781 &0.442 &0.462   \\

PETRv2$\ast$ & V2-99  &\textbf{0.856} &\textbf{0.490} &\textbf{0.463}   \\
%PETR-segonly$\ast$ & V2-99   &0.802 &0.467 &0.508   \\

\hline
\end{tabular}
\end{center}
\vspace{-1.5em}
\end{table}

\begin{table}[t]
\begin{center}
\caption{Comparison of recent 3D lane detection works on OpenLane benchmark.
PETRv2-V and PETRv2-E are our method with VoVNetV2~\cite{lee2020centermask} and EfficientNet~\cite{tan2019efficientnet} backbones.
$\ast$ is our method with 400 anchor points. The performance of Persformer is reported with EfficientNet~\cite{tan2019efficientnet} backbone.
$\ddagger$ denotes projecting 2D lane results from CondLaneNet ~\cite{liu2021condlanenet} to BEV using IPM.}
\label{lane_compare}
\setlength{\tabcolsep}{2pt}
\begin{tabular}{l|ccccc}
\hline
% \noalign{\smallskip}
Methods & F-score(\%)  & X-near & X-far & Z-near & Z-far  \\
\hline
3D-LaneNet~\cite{garnett20193d} & 44.1& 0.479 & 0.572 & 0.367 & 0.443   \\
Gen-LaneNet~\cite{guo2020gen} & 32.3 & 0.591 & 0.684 & 0.411 & 0.521 \\
Cond-IPM$\ddagger$ & 36.6 & 0.563 & 1.080 & 0.421 & 0.892 \\
PersFormer~\cite{chen2022persformer} & 50.5 & 0.485 & \textbf{0.553} & 0.364 & 0.431 \\
\hline
PETRv2-E & 51.9 & 0.493 & 0.643 & 0.322 & 0.463 \\
PETRv2-V & 57.8 & 0.427 & 0.582 & 0.293 & 0.421 \\
PETRv2-V$\ast$  & \textbf{61.2} & \textbf{0.400} & 0.573 & \textbf{0.265} & \textbf{0.413} \\

\hline
\end{tabular}
\end{center}
\vspace{-1.5em}
\end{table}

\subsection{State-of-the-art Comparison}
Tab.~\ref{table:1} compares the performance with recent works on nuScenes val set. Our method achieves state-of-the-art performance among public methods. PETRv2 achieves 39.8\% mAP and 49.4\% NDS even with ResNet-50. Tab.~\ref{table:2} shows the performance comparison on nuScenes test set. Our PETRv2 with VoVNet surpasses the PETR by a large margin (8.3\% NDS and 6.7\% mAP). Benefiting from the temporal modeling, the mAVE can achieved with 0.343m/s compared to the 0.808m/s of PETR. When compared with other temporal methods, PETRv2 surpasses the BEVDet4D~\cite{huang2022bevdet4d} with Swin-Base~\cite{liu2021swin} and BEVFormer~\cite{li2022bevformer} V2-99~\cite{lee2020centermask} by 2.2\% NDS. It shows that the temporal alignment by 3D PE can also achieve remarkable performance. It should be noted that PETRv2 can be easily employed for practical application without the explicit feature alignment. 
% The temporal 3D PE can be generated in an offline manner and served as an extra input position embedding. 

We also compare the BEV segmentation performance on nuScenes dataset. As shown in Tab.~\ref{seg_compare}, we conduct the experiments with ResNet-101 and VoV-99 backbones. Since PETRv2 is the temporal extension of PETR so we mainly compare the performance with BEVFormer for fair comparison. With ResNet-101 backbone, our PETRv2 outperforms BEVFormer on IoU-lane and IoU-Drive metrics by a large margin and achieves comparable performances on IoU-Vehicle metric. With the pretrained VoV-99 backbone, our PETRv2 achieves state-of-the-art performance.
%The results with "segonly" means training the segmentation branch only. 
% It shows that the joint learning of detection and segmentation branches will introduce the performance drop for segmentation branches. This may owe to the representation gap between these two tasks, especially for the drive and lane region. 
% For qualitative results, please refer to the visualizations in \ref{qualitive_results}.

As shown in Tab.~\ref{lane_compare}, we compare the performance with other state-of-the-art 3D lane detection methods. Since Persformer~\cite{chen2022persformer} with EfficientNet backbone is a static method, we do not use the temporal information for fair comparison. With the same EfficientNet backbone, our method achieves 51.9\% F1-score compared to the 50.5\% in Performer. With the strong pretrained VoV-99 backbone, the performance of our method is greatly improved. We also try to represent each lane with 400 anchor points and the experimental result shows that increasing the number of anchor points leads to further performance improvements. We argue that 10 anchor points are not enough to model a relatively complex 3D lane, making it difficult to make accurate prediction. It should be noted that the large number of anchor points only increase marginal computation cost in our method. The increased cost is mainly from the higher dimension of the MLP in the lane head.

% Since there are only 10 anchor points for each lane instance, it is difficult for the model to perform the regression prediction, especially at distant anchor points. Therefore, 10 anchor points are not enough to describe a relatively complex lane. 

% First, we compare the BEV segmentation results using vov backbone with the current methods. In table \ref{seg1}, we can observe that the performance of PETR is significantly higher than other methods. We also conducted a comparison of the experimental results under two settings: segmentation only and joint training. Although the performance of the method is affected in the setting of joint training, the performance of the method is still better than other methods. In addition, the setting of multi-task combined training saves a lot of calculation and training time compared with the setting of separate training. 

\begin{table*}[t!]
    \begin{center}
    % \caption{
    % The ablation studies of different components in the proposed PETRv2. PETRv2 uses two frame images as input by default. 
    % }
    \caption{The impact of 3D coordinates alignment and feature-guided position encoder. Here, CA is the 3D coordinates alignment and FPE is the proposed feature-guided position encoder.}
    \label{ablation}
    \setlength{\tabcolsep}{5pt}
    % \begin{subtable}[t]{0.94\linewidth}
        % \label{table:4c}
        \begin{tabular}{c|cc|ccccccc}
        \hline
        % \noalign{\smallskip}
         \quad\quad& \quad CA \quad &\quad FPE \quad\quad& NDS$\uparrow$\quad & mAP$\uparrow$\quad  & mATE$\downarrow$& mASE$\downarrow$& mAOE$\downarrow$& mAVE$\downarrow$ & mAAE$\downarrow$\\
        % \noalign{\smallskip}
        \hline
        % \noalign{\smallskip}
        PETR\quad& & &0.434 &0.379 &0.754 &0.272 &0.476 &0.838 &0.211\\ 
        PETR\quad& &$\checkmark$&0.449 &0.381 &0.749 &0.271 &0.462 &0.736 &0.200\\ 
        \hline
        % 1&$\checkmark$& &0.460 &0.382 &0.767 &0.270 &0.482 &0.588 &0.203\\
        PETRv2\quad& & &0.461 &0.384 &0.775 &0.270 &0.470 &0.605 &0.189\\
        % 2&$\checkmark$ &$\checkmark$&0.475 &0.391 &0.756 &0.272 &0.455 &0.531 &0.192 \\
        PETRv2\quad&\quad$\checkmark$& &0.482 &0.393 &0.774 &0.272 &0.486 &0.429 &0.187 \\
        PETRv2\quad&\quad$\checkmark$ &$\checkmark$ &\textbf{0.496} &\textbf{0.401} &\textbf{0.745} &\textbf{0.268} &\textbf{0.448} &\textbf{0.394} &\textbf{0.184}\\
        \hline
        \end{tabular}

    \label{tab:ablation}
\end{center}
\vspace{-1.0em}
\end{table*}

\subsection{Ablation Study}
In this section, we conduct the ablations with VoVNet-99 backbone. The backbone is pretrained on DDAM15M dataset~\cite{park2021dd3d} and train set of Nuscenes~\cite{caesar2020nuscenes}. The input image size is 800 $\times$ 320 and the model is trained with 24 epochs. The number of detection queries is set to 900.
% The model can be trained on 8 2080ti GPU within 24 hours.

% \noindent \textbf{Temporal Modeling.}
Here we explore the effect of two key components in our design: 3D coordinates alignment (CA) and feature-guided position encoder (FPE). For the ablation study, we only trained the 3D detection branch for clarity. As shown in Tab.~\ref{ablation}(a), without CA, PETRv2 only improves the performance by 2.7\% NDS and 0.5\% mAP. With CA, the performance is further improved by 2.1\% NDS and 0.9\% mAP. The mAVE metric is decreased to 0.429 m/s, which shows a large margin compared to the original PETR baseline. To verify the effectiveness of FPE, we replace the 3D position encoder in PETR with FPE. The NDS metric is increased by 1.5\% while mAP is only increased by 0.2\%. When we apply the FPE on PETRv2, the mAP achieves a relatively higher improvement (0.8\%). It indicates that FPE module is also beneficial to the temporal version of PETR.

\subsection{Robustness analysis}
% Robustness to extrinsic noise
% table
% 1. extrinsic noise: setting, values
Tab.~\ref{table:rob_rot} reports a summary of quantitative results on the nuScenes dataset with extrinsics noises during inference. We compare PETRv2, PETR and PETR + FPE (FPE denotes the feature-guided position encoder).
As the noise increases, the performance of all three models decreases continually, indicating the impact of extrinsics noises. In the extreme noise setting $R_{max}=8$, PETRv2 drops 4.12\% mAP and 2.85\% NDS, PETR+FPE drops 4.68\% mAP and 3.42\% NDS, while PETR drops 6.33\% mAP and 4.54\% NDS. We observe that FPE improves the robustness to extrinsics noises, while temporal extension with multiple frames does not bring significant robustness gains.

% mf extr noise: rot+trans
\begin{table}[h]
    % \begin{center}
    \caption{Quantitative results on the nuScenes val set with extrinsics noises. The metrics in each cell are mAP[\%]. $R_{max}=M$ denotes the maximum angle of three axes is M in degree. }
    \label{table:rob_rot}
    \resizebox{0.5\textwidth}{!}{
    \setlength{\tabcolsep}{5pt}
    \begin{centering}
    \begin{tabular}{c|c|c|c}
        \hline
        % \noalign{\smallskip}
        Methods & $R_{max}=2$ & $R_{max}=4$ & $R_{max}=6$\\% &  $R_{max}=8$    Original & 
        % \noalign{\smallskip}
        \hline
        % \noalign{\smallskip}
        PETR    & 36.71 ($\downarrow$1.16) & 34.58 ($\downarrow$3.29) & 32.79 ($\downarrow$5.08) \\ %& 37.87 
        PETR+FPE & 37.17 ($\downarrow$0.96) & 35.83 ($\downarrow$2.30) & 34.47 ($\downarrow$3.66) \\ %  & 38.13
        PETRv2 & 39.13 ($\downarrow$0.95) & 37.69 ($\downarrow$2.15) & 36.66 ($\downarrow$3.42) \\ % & 40.08 
        \hline
        \end{tabular}
    \end{centering}
    }
    % \end{center}
\end{table}

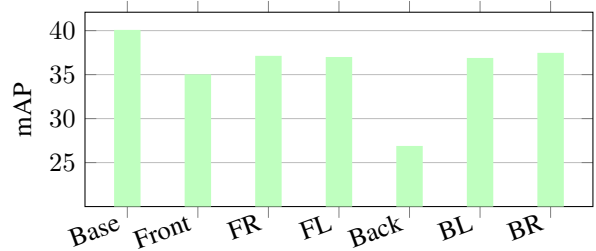
\begin{figure}[h]
  \begin{tikzpicture}
    \begin{axis}[
        width=\linewidth,
        height=0.5\linewidth,
        ybar,
        ylabel=mAP,
        ylabel near ticks,
        ymajorgrids=true,
        ytick={25, 30, 35, 40, 45, 50},
        ymin=20,
        xtick={0,...,6},
        xticklabels={Base, Front, FR, FL, Back, BL, BR},
        x tick label style={rotate=20, anchor=east},
      ]
      \addplot[draw=none, fill=green!25]
      coordinates {
          (0, 40.08)
          (1, 35.03)
          (2, 37.13)
          (3, 37.00)
          (4, 26.89)
          (5, 36.89)
          (6, 37.47)
        };
    \end{axis}
  \end{tikzpicture}
  \caption{The performance on nuScenes val when losing each of camera images. FR, FL, BL and BR denote the front-right, front-left, back-left and back-right, respectively.}
  \label{fig:ab_cam_miss}
\end{figure}

We also show how the model performs when randomly losing one camera in Fig.~\ref{fig:ab_cam_miss}. Among these six cameras of nuScenes dataset, the front and back cameras are the most important ones, and their absences leads to a drop of 5.05\% and 13.19\% mAP, respectively. The back camera is especially essential due to its large field of view ($120^{\circ}$). Losing other cameras also brings an average performance decrease of 2.93\% mAP and 1.93\% NDS. Note that the overlap region between cameras is very small for the nuScenes dataset, thus performance drop caused by any camera miss is hard to be compensated by adjacent ones. In practice, sensor redundancy is necessary in case of emergency and complementary of cameras requires deeper explorations.

The effect of camera time delay is demonstrated in Tab.~\ref{table:rob_time_delay}. In nuScenes, key frames are annotated with ground-truth, and we leverage unannotated frames between key frames as input images to simulate the time delay. 
The delay of 0.083s leads to a drop of 3.19\% mAP and 8.4\% NDS, indicating the significant impact of time delay. When time delay increase to over 0.3s, the performance sharply decreases to 26.08\% mAP and 36.54\% NDS.
Since time delay is inevitable in real-world systems and affects detection a lot, more attention is supposed to pay to it.

\begin{table}[h]
    \caption{The performance impact (on mAP metric) of camera time delay. Here, the time delay unit $T \approx 0.083$s.}
    \label{table:rob_time_delay}
    \resizebox{0.5\textwidth}{!}{
    \setlength{\tabcolsep}{5pt}
    \begin{centering}
    \begin{tabular}{c|c|c|c}
        \hline
        % \noalign{\smallskip}
        Time delay & T & 2T & 3T\\  % & Original 
        % \noalign{\smallskip}
        \hline
        % \noalign{\smallskip}
        PETRv2  & 36.89 ($\downarrow$3.19) & 33.99 ($\downarrow$6.09) & 30.91 ($\downarrow$9.17) \\%& 26.08 ($\downarrow$14.00) & 40.08 
        \hline
        \end{tabular}
    \end{centering}
    }
% \vspace{-5pt}
\end{table}

\section{Conclusion}
In this paper, we introduce PETRv2, a unified framework for 3D perception from multi-camera images. PETRv2 extends the PETR baseline with temporal modeling and multi-task learning. With the temporal alignment on 3D position embedding, PETRv2 naturally achieves the multi-frame modeling and improves the 3D detection performance.
% PETRv2 also supports the multi-task learning, such as BEV segmentation, by adding a set of task-specific queries. 
For a fully understanding of PETRv2 framework, we further provide a detailed analysis on the robustness of PETRv2 under three types of simulated sensor errors. We hope PETRv2 can serve as a strong baseline and a unified framework for 3D perception. In the near future, we amy explore large-scale pretraining, more 3D vision tasks and multi-modal fusion for autonomous driving system.

%%%%%%%%% REFERENCES
{\small
\bibliographystyle{ieee_fullname}
\bibliography{egbib} 

\begin{thebibliography}{10}\itemsep=-1pt

\bibitem{bai2022curveformer}
Yifeng Bai, Zhirong Chen, Zhangjie Fu, Lang Peng, Pengpeng Liang, and Erkang
  Cheng.
\newblock Curveformer: 3d lane detection by curve propagation with curve
  queries and attention.
\newblock {\em arXiv preprint arXiv:2209.07989}, 2022.

\bibitem{brazil2019m3d}
Garrick Brazil and Xiaoming Liu.
\newblock M3d-rpn: Monocular 3d region proposal network for object detection.
\newblock In {\em Proceedings of the IEEE/CVF International Conference on
  Computer Vision}, pages 9287--9296, 2019.

\bibitem{caesar2020nuscenes}
Holger Caesar, Varun Bankiti, Alex~H Lang, Sourabh Vora, Venice~Erin Liong,
  Qiang Xu, Anush Krishnan, Yu Pan, Giancarlo Baldan, and Oscar Beijbom.
\newblock nuscenes: A multimodal dataset for autonomous driving.
\newblock In {\em Proceedings of the IEEE/CVF conference on computer vision and
  pattern recognition}, pages 11621--11631, 2020.

\bibitem{carion2020detr}
Nicolas Carion, Francisco Massa, Gabriel Synnaeve, Nicolas Usunier, Alexander
  Kirillov, and Sergey Zagoruyko.
\newblock End-to-end object detection with transformers.
\newblock In {\em European conference on computer vision}, pages 213--229.
  Springer, 2020.

\bibitem{chen2022persformer}
Li Chen, Chonghao Sima, Yang Li, Zehan Zheng, Jiajie Xu, Xiangwei Geng,
  Hongyang Li, Conghui He, Jianping Shi, Yu Qiao, et~al.
\newblock Persformer: 3d lane detection via perspective transformer and the
  openlane benchmark.
\newblock {\em arXiv preprint arXiv:2203.11089}, 2022.

\bibitem{chen2016monocular}
Xiaozhi Chen, Kaustav Kundu, Ziyu Zhang, Huimin Ma, Sanja Fidler, and Raquel
  Urtasun.
\newblock Monocular 3d object detection for autonomous driving.
\newblock In {\em Proceedings of the IEEE conference on computer vision and
  pattern recognition}, pages 2147--2156, 2016.

\bibitem{garnett20193d}
Noa Garnett, Rafi Cohen, Tomer Pe'er, Roee Lahav, and Dan Levi.
\newblock 3d-lanenet: end-to-end 3d multiple lane detection.
\newblock In {\em Proceedings of the IEEE/CVF International Conference on
  Computer Vision}, pages 2921--2930, 2019.

\bibitem{guo2020gen}
Yuliang Guo, Guang Chen, Peitao Zhao, Weide Zhang, Jinghao Miao, Jingao Wang,
  and Tae~Eun Choe.
\newblock Gen-lanenet: A generalized and scalable approach for 3d lane
  detection.
\newblock In {\em European Conference on Computer Vision}, pages 666--681.
  Springer, 2020.

\bibitem{he2016resnet}
Kaiming He, Xiangyu Zhang, Shaoqing Ren, and Jian Sun.
\newblock Deep residual learning for image recognition.
\newblock In {\em Proceedings of the IEEE conference on computer vision and
  pattern recognition}, pages 770--778, 2016.

\bibitem{hu2021fiery}
Anthony Hu, Zak Murez, Nikhil Mohan, Sof{\'\i}a Dudas, Jeffrey Hawke, Vijay
  Badrinarayanan, Roberto Cipolla, and Alex Kendall.
\newblock Fiery: Future instance prediction in bird's-eye view from surround
  monocular cameras.
\newblock In {\em Proceedings of the IEEE/CVF International Conference on
  Computer Vision}, pages 15273--15282, 2021.

\bibitem{huang2022bevdet4d}
Junjie Huang and Guan Huang.
\newblock Bevdet4d: Exploit temporal cues in multi-camera 3d object detection.
\newblock {\em arXiv preprint arXiv:/2203.17054}, 2021.

\bibitem{huang2021bevdet}
Junjie Huang, Guan Huang, Zheng Zhu, and Dalong Du.
\newblock Bevdet: High-performance multi-camera 3d object detection in
  bird-eye-view.
\newblock {\em arXiv preprint arXiv:2112.11790}, 2021.

\bibitem{jorgensen2019monocular}
Eskil J{\"o}rgensen, Christopher Zach, and Fredrik Kahl.
\newblock Monocular 3d object detection and box fitting trained end-to-end
  using intersection-over-union loss.
\newblock {\em arXiv preprint arXiv:1906.08070}, 2019.

\bibitem{kehl2017ssd}
Wadim Kehl, Fabian Manhardt, Federico Tombari, Slobodan Ilic, and Nassir Navab.
\newblock Ssd-6d: Making rgb-based 3d detection and 6d pose estimation great
  again.
\newblock In {\em Proceedings of the IEEE international conference on computer
  vision}, pages 1521--1529, 2017.

\bibitem{ku2019monocular}
Jason Ku, Alex~D Pon, and Steven~L Waslander.
\newblock Monocular 3d object detection leveraging accurate proposals and shape
  reconstruction.
\newblock In {\em Proceedings of the IEEE/CVF conference on computer vision and
  pattern recognition}, pages 11867--11876, 2019.

\bibitem{kuhn1955hungarian}
Harold~W Kuhn.
\newblock The hungarian method for the assignment problem.
\newblock {\em Naval research logistics quarterly}, 2(1-2):83--97, 1955.

\bibitem{lee2020centermask}
Youngwan Lee and Jongyoul Park.
\newblock Centermask: Real-time anchor-free instance segmentation.
\newblock In {\em Proceedings of the IEEE/CVF conference on computer vision and
  pattern recognition}, pages 13906--13915, 2020.

\bibitem{li2022dn}
Feng Li, Hao Zhang, Shilong Liu, Jian Guo, Lionel~M Ni, and Lei Zhang.
\newblock Dn-detr: Accelerate detr training by introducing query denoising.
\newblock In {\em Proceedings of the IEEE/CVF Conference on Computer Vision and
  Pattern Recognition}, pages 13619--13627, 2022.

\bibitem{li2021hdmapnet}
Qi Li, Yue Wang, Yilun Wang, and Hang Zhao.
\newblock Hdmapnet: A local semantic map learning and evaluation framework.
\newblock {\em arXiv preprint arXiv:2107.06307}, 2021.

\bibitem{li2022bevformer}
Zhiqi Li, Wenhai Wang, Hongyang Li, Enze Xie, Chonghao Sima, Tong Lu, Qiao Yu,
  and Jifeng Dai.
\newblock Bevformer: Learning bird's-eye-view representation from multi-camera
  images via spatiotemporal transformers.
\newblock {\em arXiv preprint arXiv:2203.17270}, 2022.

\bibitem{lin2017focal}
Tsung-Yi Lin, Priya Goyal, Ross Girshick, Kaiming He, and Piotr Doll{\'a}r.
\newblock Focal loss for dense object detection.
\newblock In {\em Proceedings of the IEEE international conference on computer
  vision}, pages 2980--2988, 2017.

\bibitem{liu2021condlanenet}
Lizhe Liu, Xiaohao Chen, Siyu Zhu, and Ping Tan.
\newblock Condlanenet: a top-to-down lane detection framework based on
  conditional convolution.
\newblock In {\em Proceedings of the IEEE/CVF International Conference on
  Computer Vision}, pages 3773--3782, 2021.

\bibitem{liu2022dab}
Shilong Liu, Feng Li, Hao Zhang, Xiao Yang, Xianbiao Qi, Hang Su, Jun Zhu, and
  Lei Zhang.
\newblock Dab-detr: Dynamic anchor boxes are better queries for detr.
\newblock {\em arXiv preprint arXiv:2201.12329}, 2022.

\bibitem{liu2022petr}
Yingfei Liu, Tiancai Wang, Xiangyu Zhang, and Jian Sun.
\newblock Petr: Position embedding transformation for multi-view 3d object
  detection.
\newblock {\em arXiv preprint arXiv:2203.05625}, 2022.

\bibitem{liu2021swin}
Ze Liu, Yutong Lin, Yue Cao, Han Hu, Yixuan Wei, Zheng Zhang, Stephen Lin, and
  Baining Guo.
\newblock Swin transformer: Hierarchical vision transformer using shifted
  windows.
\newblock In {\em Proceedings of the IEEE/CVF International Conference on
  Computer Vision}, pages 10012--10022, 2021.

\bibitem{loshchilov2016sgdr}
Ilya Loshchilov and Frank Hutter.
\newblock Sgdr: Stochastic gradient descent with warm restarts.
\newblock {\em arXiv preprint arXiv:1608.03983}, 2016.

\bibitem{loshchilov2017decoupled}
Ilya Loshchilov and Frank Hutter.
\newblock Decoupled weight decay regularization.
\newblock {\em arXiv preprint arXiv:1711.05101}, 2017.

\bibitem{meng2021conditional}
Depu Meng, Xiaokang Chen, Zejia Fan, Gang Zeng, Houqiang Li, Yuhui Yuan, Lei
  Sun, and Jingdong Wang.
\newblock Conditional detr for fast training convergence.
\newblock In {\em Proceedings of the IEEE/CVF International Conference on
  Computer Vision}, pages 3651--3660, 2021.

\bibitem{mousavian20173d}
Arsalan Mousavian, Dragomir Anguelov, John Flynn, and Jana Kosecka.
\newblock 3d bounding box estimation using deep learning and geometry.
\newblock In {\em Proceedings of the IEEE conference on Computer Vision and
  Pattern Recognition}, pages 7074--7082, 2017.

\bibitem{pan2020cross}
Bowen Pan, Jiankai Sun, Ho~Yin~Tiga Leung, Alex Andonian, and Bolei Zhou.
\newblock Cross-view semantic segmentation for sensing surroundings.
\newblock {\em IEEE Robotics and Automation Letters}, 5(3):4867--4873, 2020.

\bibitem{park2021dd3d}
Dennis Park, Rares Ambrus, Vitor Guizilini, Jie Li, and Adrien Gaidon.
\newblock Is pseudo-lidar needed for monocular 3d object detection?
\newblock In {\em Proceedings of the IEEE/CVF International Conference on
  Computer Vision}, pages 3142--3152, 2021.

\bibitem{peng2022bevsegformer}
Lang Peng, Zhirong Chen, Zhangjie Fu, Pengpeng Liang, and Erkang Cheng.
\newblock Bevsegformer: Bird's eye view semantic segmentation from arbitrary
  camera rigs.
\newblock {\em arXiv preprint arXiv:2203.04050}, 2022.

\bibitem{philion2020lift}
Jonah Philion and Sanja Fidler.
\newblock Lift, splat, shoot: Encoding images from arbitrary camera rigs by
  implicitly unprojecting to 3d.
\newblock In {\em European Conference on Computer Vision}, pages 194--210.
  Springer, 2020.

\bibitem{rukhovich2022imvoxelnet}
Danila Rukhovich, Anna Vorontsova, and Anton Konushin.
\newblock Imvoxelnet: Image to voxels projection for monocular and multi-view
  general-purpose 3d object detection.
\newblock In {\em Proceedings of the IEEE/CVF Winter Conference on Applications
  of Computer Vision}, pages 2397--2406, 2022.

\bibitem{simonelli2019disentangling}
Andrea Simonelli, Samuel~Rota Bulo, Lorenzo Porzi, Manuel L{\'o}pez-Antequera,
  and Peter Kontschieder.
\newblock Disentangling monocular 3d object detection.
\newblock In {\em Proceedings of the IEEE/CVF International Conference on
  Computer Vision}, pages 1991--1999, 2019.

\bibitem{tan2019efficientnet}
Mingxing Tan and Quoc Le.
\newblock Efficientnet: Rethinking model scaling for convolutional neural
  networks.
\newblock In {\em International conference on machine learning}, pages
  6105--6114. PMLR, 2019.

\bibitem{wang2022pgd}
Tai Wang, ZHU Xinge, Jiangmiao Pang, and Dahua Lin.
\newblock Probabilistic and geometric depth: Detecting objects in perspective.
\newblock In {\em Conference on Robot Learning}, pages 1475--1485. PMLR, 2022.

\bibitem{wang2021fcos3d}
Tai Wang, Xinge Zhu, Jiangmiao Pang, and Dahua Lin.
\newblock Fcos3d: Fully convolutional one-stage monocular 3d object detection.
\newblock In {\em Proceedings of the IEEE/CVF International Conference on
  Computer Vision}, pages 913--922, 2021.

\bibitem{wang2022detr3d}
Yue Wang, Guizilini Vitor~Campagnolo, Tianyuan Zhang, Hang Zhao, and Justin
  Solomon.
\newblock Detr3d: 3d object detection from multi-view images via 3d-to-2d
  queries.
\newblock In {\em In Conference on Robot Learning}, pages 180--191, 2022.

\bibitem{xie2022m}
Enze Xie, Zhiding Yu, Daquan Zhou, Jonah Philion, Anima Anandkumar, Sanja
  Fidler, Ping Luo, and Jose~M Alvarez.
\newblock M\^{} 2bev: Multi-camera joint 3d detection and segmentation with
  unified birds-eye view representation.
\newblock {\em arXiv preprint arXiv:2204.05088}, 2022.

\bibitem{xie2017aggregated}
Saining Xie, Ross Girshick, Piotr Doll{\'a}r, Zhuowen Tu, and Kaiming He.
\newblock Aggregated residual transformations for deep neural networks.
\newblock In {\em Proceedings of the IEEE conference on computer vision and
  pattern recognition}, pages 1492--1500, 2017.

\bibitem{yin2021center}
Tianwei Yin, Xingyi Zhou, and Philipp Krahenbuhl.
\newblock Center-based 3d object detection and tracking.
\newblock In {\em Proceedings of the IEEE/CVF conference on computer vision and
  pattern recognition}, pages 11784--11793, 2021.

\bibitem{zhou2022cross}
Brady Zhou and Philipp Kr{\"a}henb{\"u}hl.
\newblock Cross-view transformers for real-time map-view semantic segmentation.
\newblock {\em arXiv preprint arXiv:2205.02833}, 2022.

\bibitem{zhou2019objects}
Xingyi Zhou, Dequan Wang, and Philipp Kr{\"a}henb{\"u}hl.
\newblock Objects as points.
\newblock {\em arXiv preprint arXiv:1904.07850}, 2019.

\bibitem{zhu2019class}
Benjin Zhu, Zhengkai Jiang, Xiangxin Zhou, Zeming Li, and Gang Yu.
\newblock Class-balanced grouping and sampling for point cloud 3d object
  detection.
\newblock {\em arXiv preprint arXiv:1908.09492}, 2019.

\bibitem{zhu2020deformable}
Xizhou Zhu, Weijie Su, Lewei Lu, Bin Li, Xiaogang Wang, and Jifeng Dai.
\newblock Deformable detr: Deformable transformers for end-to-end object
  detection.
\newblock {\em arXiv preprint arXiv:2010.04159}, 2020.

\end{thebibliography}
}

\end{document}